\documentclass[10pt,twocolumn,letterpaper]{article}

\usepackage{cvpr}              
\usepackage{graphicx}
\usepackage{amsmath}
\usepackage{amssymb}
\usepackage{booktabs}
\usepackage{algorithm}
\usepackage[linesnumbered,ruled,lined,algo2e]{algorithm2e}
\usepackage{multirow}
\usepackage{colortbl}
\usepackage[dvipsnames]{xcolor}

\definecolor{cvprblue}{rgb}{0.21,0.49,0.74}
\usepackage[pagebackref,breaklinks,colorlinks,citecolor=cvprblue]{hyperref}

\def\paperID{8228} 
\def\confName{CVPR}
\def\confYear{2024}

\title{ExACT: Language-guided Conceptual Reasoning and Uncertainty Estimation for Event-based Action Recognition and More}

\author{Jiazhou Zhou$^{1}$ \quad Xu Zheng$^{1}$ \quad Yuanhuiyi Lyu$^{1}$ \quad Lin Wang$^{1}$$^{,2}$\thanks{Corresponding author.}\\
$^{1}$AI Thrust, HKUST(GZ) \quad $^{2}$Dept. of CSE, HKUST\\
{\tt\small jiazhouzhou@hkust-gz.edu.cn, yuanhuiyilv@hkust-gz.edu.cn, zhengxu128@gmail.com, linwang@ust.hk}
\\
\small{Project Page: \url{https://vlislab22.github.io/ExACT/}}}

\begin{document}
\maketitle
\begin{abstract}
    Event cameras have recently been shown beneficial for practical vision tasks, such as action recognition, thanks to their high temporal resolution, power efficiency, and reduced privacy concerns. 
    However, current research is hindered by 1) the difficulty in processing events because of their prolonged duration and dynamic actions with complex and ambiguous semantics and 2) the redundant action depiction of the event frame representation with fixed stacks.  
    We find language naturally conveys abundant semantic information, rendering it stunningly superior in reducing semantic uncertainty.  In light of this, we propose ExACT, a novel approach that, for the first time, 
    tackles event-based action recognition from a cross-modal conceptualizing perspective. Our ExACT brings two technical contributions. Firstly, we propose an adaptive fine-grained event (AFE) representation to adaptively filter out the repeated events for the stationary objects while preserving dynamic ones. This subtly enhances the performance of ExACT without extra computational cost. Then, we propose a conceptual reasoning-based uncertainty estimation module, which simulates the recognition process to enrich the semantic representation. In particular, conceptual reasoning builds the temporal relation based on the action semantics, and uncertainty estimation tackles the semantic uncertainty of actions based on the distributional representation. 
    Experiments show that our ExACT achieves superior recognition accuracy of 94.83\%(+2.23\%), 90.10\%(+37.47\%) and 67.24\% on PAF, HARDVS and our SeAct datasets respectively.
\end{abstract}

\begin{figure}[t!]
\centering
\includegraphics[width=1.0\columnwidth]
{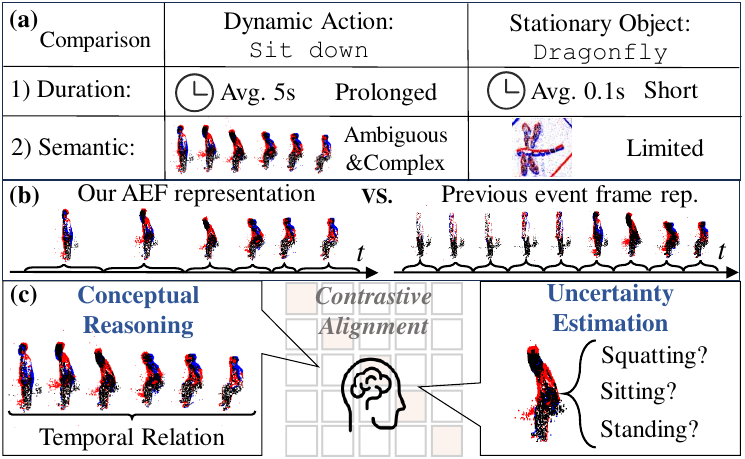}
\vspace{-16pt}
\caption{\textbf{(a)} Unlike stationary objects, \eg, {\fontfamily{qcr}\selectfont‘Dragonfly'} with short duration (0.1s) and limited semantics, dynamic actions like {\fontfamily{qcr}\selectfont‘Sit down'} have the prolonged duration (5s) with ambiguous and complex semantics. 
\textbf{(b)} Compared with previous event representation, stacking events with fixed counts, we adaptively filter out events recording stationary actions while preserving dynamic ones; 
\textbf{(c)} We introduce language guidance to stimulate the recognition process, particularly focusing on conceptually reasoning temporal relations and estimating uncertain semantics.}
\label{Teaser}
\vspace{-16pt}
\end{figure}

\vspace{-15pt}
\section{Introduction}
\label{sec:intro}

Action recognition is a crucial vision task with many applications, such as robot navigation~\cite{sun2022human,mavrogiannis2023core}, and abnormal human behavior recognition ~\cite{lentzas2020non,pareek2021survey}. Many frame-based learning approaches have been presented, leading to impressive performance improvements~\cite{kong2022human,sun2022human}. 
However, these methods may not be ideal solutions for power-constrained scenarios, \eg, surveillance~\cite{kong2022human,beddiar2020vision}. 
RGB cameras also degrade due to environment bias~\cite{plizzari2022e2} like motion blur and lighting variations. Moreover, frame-based cameras trigger considerable privacy concerns as they directly capture users' appearance.

Recently, bio-inspired event cameras are gaining popularity~\cite{gao2023action,liu2021event,amir2017low}, which ignore the background and only record moving objects.
This leads to sensing efficiency and resilience to rapid motion and illumination changes with low power consumption.
Also, event cameras mostly reflect objects' edges, which alleviates users' privacy concerns like skin color and gender.  Owing to these advantages, event-based action recognition offers more pragmatic solutions for real-world implementations.
This has inspired research endeavors~\cite{gao2023action,sabater2022event,xie2023event,liu2021event,xie2022vmv,wang2023sstformer,plizzari2022e2} in event-based action recognition areas with plausible performance.

However, the above methods are deficient for two reasons: 1) They have a limited capacity for recognizing a large number of different human actions, as demonstrated by experiments on the HARDVS~\cite{wang2022hardvs} dataset with 300 categories in Tab.~\ref{result_recognition}. This probably arises from the complex and ambiguous semantics induced by the dynamic actions and prolonged duration (around 5s \cite{wang2022hardvs}), compared to objects with short duration (around 0.1s \cite{wang2022hardvs}) and limited semantics. 
For example, as shown in Fig.~\ref{Teaser} (a), {\fontfamily{qcr}\selectfont‘Dragonfly'} \textit{vs.} {\fontfamily{qcr}\selectfont‘Sit down'}. 
2) The lack of \textit{tailored event representation} as the raw events are directly stacked into event frames with fixed stacks, resulting in event frames with either overlapped or vague edge information depicting the same action, see Fig.~\ref{Case_study} and Fig.~\ref{Teaser} (b).




Recent advancements in vision-language models (VLMs) \cite{CLIP,chen2023vlp,du2022survey} have pioneered ideas that incorporate semantic concepts across text and vision modalities, aiming at simulating the human processes of conceptualization and reasoning~\cite{su2021self,sun2020view,ji2023map}.
\textit{The key insight is that language naturally conveys inherent semantic richness, which can be beneficial for modeling semantic uncertainty and establishing complex semantic relations}.
Inspired by this, we introduce language as guidance for event-based action recognition. As the first exploration, the research hurdles include: 1) How to represent events to depict dynamic actions in detail without redundant event frames? 2) How to integrate text embeddings with event embeddings to help reason complex semantics of dynamic actions and reduce semantic uncertainty? 


To this end, we propose a novel ExACT framework to tackle event-based action recognition from a cross-modal conceptual reasoning perspective, as depicted in Fig.~\ref{Teaser} (b) and (c). 
To address the first challenge, an Adaptive Fine-grained Event (AFE) representation (Sec.~\ref{sec:representation}) is inspired by the `overlapped action regions' observed in Fig.~\ref{Case_study}.
These regions indicate an excessive stack of events in one frame, which is inevitable for previous frame-based representations with fixed stacks. 
Differently, our AFE recursively and offline find the dividing line of different actions based on the overlapped regions. It eliminates repeated events and preserves dynamic actions, thus enhancing model performance without extra computational costs (Tab.~\ref{result_representation}).



For the second challenge, we propose a novel Conceptual Reasoning-based Uncertainty Estimation (CRUE) module (Sec.~\ref{Conceptualizing}) to simulate the action recognition process of humans. Concretely, CRUE initially establishes the temporal relation of event frames by leveraging the text embeddings to reason each frame's semantics and then obtain the fused event embeddings. Subsequently, CRUE converts event and text embeddings from discrete representation to distributional representation, where the distribution variance quantifies semantic uncertainty. In this way, our proposed CRUE module establishes a semantic-abundant and uncertainty-aware embedding space to enhance model performance (Tab.~\ref{ab_conceptualizing}).

\begin{figure*}[t]
\centering
\includegraphics[width=1.0\textwidth]{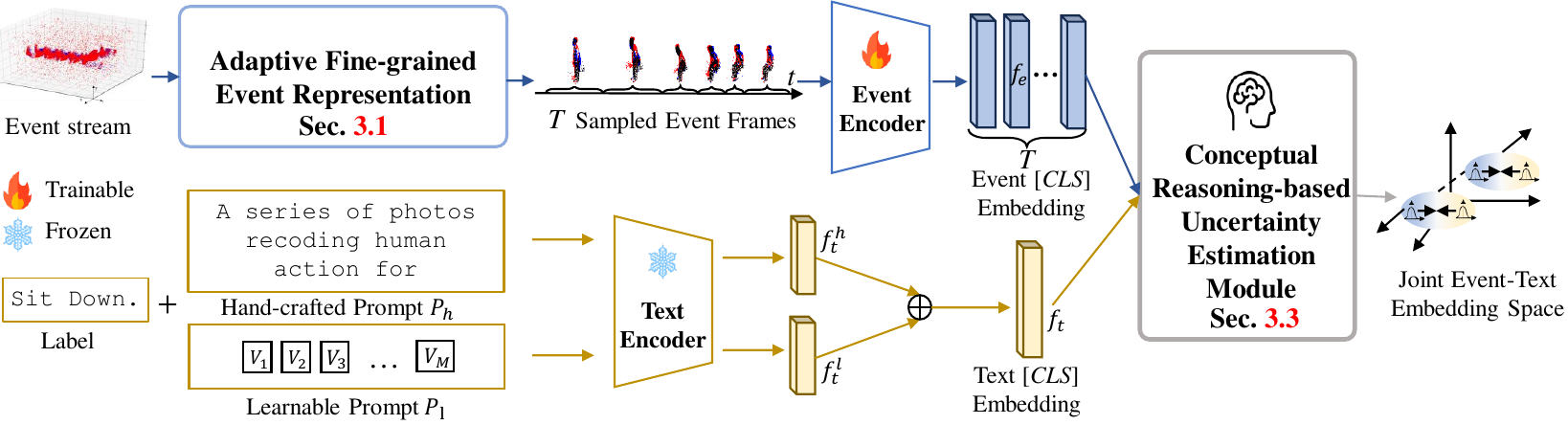}
\vspace{-15pt}
\caption{\textbf{Overall framework of our proposed ExACT framework.} 
It consists of four components: 
(1) the AFE representation recursively eliminates repeated events and generates event frames depicting dynamic actions (Sec.~\ref{sec:representation});
(2) the event encoder and (3) the text encoder, responsible for the event and text embedding, respectively (Sec.~\ref{sec:feature_embedding}); (4) the CRUE module simulates the action recognition process to establish the complex semantic relations for sub-actions and reduce the semantic uncertainty. (Sec.~\ref{sec:CRUE})}\label{Framework}
\vspace{-10pt}
\end{figure*}

Meanwhile, as existing datasets solely provide category-level labels, we propose the SeAct dataset, consisting of 58 categories of actions, with semantic-abundant caption-level labels. Our dataset serves as the first dataset for event-text action recognition (67.24\% accuracy). 
We also conduct extensive experiments to show that our ExACT outperforms previous methods~, \eg,~\cite{lin2019tsm, gao2023action} on the public datasets, PAF~\cite{miao2019neuromorphic} 94.83\% accuracy (+2.23\%) and especially on HARDVS~\cite{wang2022hardvs} 90.10\% accuracy (+37.47\%) by a large margin.
Beyond action recognition, our ExACT can be flexibly applied to the event-text retrieval task.

In summary, our main contributions are:
(I) We propose the ExACT-- the first framework utilizing language guidance for event-based action recognition; 
(II) We propose the CRUE module to mimic human action recognition, creating a rich, uncertainty-aware cross-modal embedding space for action recognition. Also, our AFE representation adaptively filters redundant events, yielding effective representation for dynamic actions.
(III) We introduce the SeAct dataset with detailed text captions for evaluating the recognizing ability of actions composed of multiple sub-actions with different semantics. Extensive experiments demonstrate the superiority of our ExACT framework on our SeAct dataset and public datasets.

\vspace{-7pt}
\section{Related Works}
\vspace{-2pt}
\label{sec:related}
\noindent \textbf{Event-based Action Recognition} methods can be categorized into the event-only and event-other-modality methods. For event-only frameworks, the most widespread techniques involve stacking the event stream into compact frames, followed by the utilization of off-the-shelf Convolutional Neural Networks (CNNs)~\cite{gao2023action} or Vision Transformers (ViTs)~\cite{sabater2022event, xie2023event} for effective feature extraction. This approach currently exhibits state-of-the-art performance owing to the excellent CNN/ViT backbones performance. 
Meanwhile, considering the asynchronous characteristics of event data, the research community has explored the applicability of bio-inspired Spiking Neural Networks (SNNs)~\cite{liu2021event} and the spatiotemporal capabilities of Graph Convolutional Networks (GCNs)~\cite{xie2022vmv} to more aptly resonate with the unique structure of event data. However, these approaches have yielded suboptimal performance and exhibited limited adaptability, partly due to the specialized hardware requisites inherent to SNNs.

Concurrently, there have been endeavors to integrate event data with additional modalities. For example, incorporating the abundance of color and texture information in RGB~\cite{wang2023sstformer} data with event information or utilizing the supplementary motion knowledge in optical flow~\cite{plizzari2022e2}. In summary, most of these approaches rely on dense consecutive event frames, inevitably resulting in redundant frames with overlapped actions and uncertain semantics. To represent events to depict detailed dynamic actions, the AFE representation is proposed to sample event frames adaptively without introducing extra computation costs.

\noindent \textbf{Vision-laguage Models (VLMs)}
Recently, there has been a growing interest in large-scale pre-trained VLMs~\cite{du2022survey, chen2023vlp} for multimodal representations. 
Inspired by it, several pioneering works~\cite{zhou2023clip,wu2023eventclip,cho2023label} have investigated the potential for transferring VLMs' capability to the event modality, thereby revitalizing the best performance of objection recognition. In addition, the remarkable zero-shot capability of VLMs has motivated researchers to explore event-based label-free~\cite{cho2023label} or zero (few)-shot applications~\cite{zhou2023clip}, thus addressing the scarcity of high-quality event datasets. Nevertheless, prior event-text methods focus on recognizing objects with limited semantics but fail to recognize events recording actions with prolonged time duration and complex and ambiguous semantics.
Therefore, Our ExACT aims at enhancing event-based action recognition from a cross-modal conceptual reasoning perspective.

\vspace{-13pt}
\section{The Proposed ExACT Framework}
\noindent\textbf{Overview.}
An overview of our ExACT framework is depicted in Fig.~\ref{Framework}.
\textit{The key idea of ExACT is to introduce language as guidance for estimating semantic uncertainty and establishing semantic relations for event-based action recognition.} 
The following subsections explain the technical details of 1) the proposed Adaptive Fine-grained Event (AFE) representation (Sec.~\ref{sec:representation}); 2) the event encoder and the text encoder (Sec.~\ref{sec:feature_embedding}); 3) the Conceptual Reasoning-based Uncertainty
Estimation (CRUE) module (Sec.~\ref{sec:CRUE}). Besides, in Sec.~\ref{sec:SemanticAct}, we introduce our proposed \textbf{s}emantic-abundant \textbf{e}vent-based \textbf{act}ion recognition (\textbf{SeAct}) dataset as the first dataset for event-text action recognition.



\begin{figure}[t!]
\centering
\includegraphics[width=0.95\columnwidth]{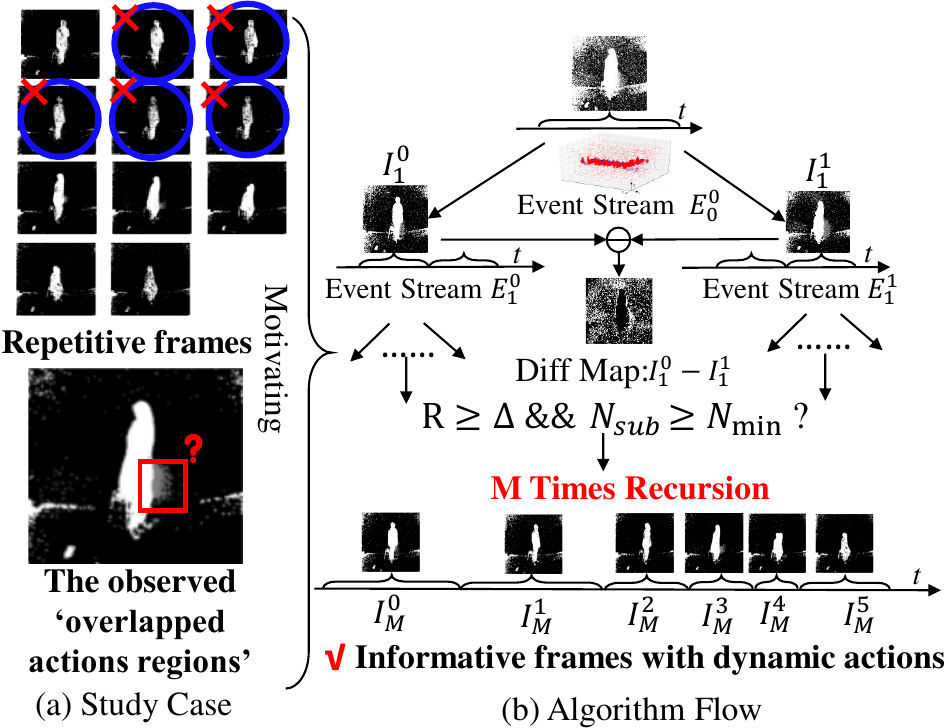}
\vspace{-4pt}
\caption{
\textbf{(a)} Unlike existing methods often lead to repetitive event frames, our AFE representation adaptively filters out repetitive events for the same action based on the observed overlapped action regions; \textbf{(b)} Illustration of the AFE representation. 
}\label{Case_study}
\vspace{-12pt}
\end{figure}

\subsection{The AFE Representation}
\label{sec:representation}
Most event-based action recognition models~\cite{gao2023action, sabater2022event, xie2023event} primarily rely on event frame representations~\cite{xu2021motion,manderscheid2019speed}, which is compatible with off-the-shelf CNN/ViT backbones. For these models, the event stream is spatially integrated into frames with fixed event counts or time duration~\cite{zheng2023deep}, as shown in Fig.~\ref{Teaser} (b). However, the high temporal resolution of event data inevitably leads to a profusion of repetitive event frames displaying the same stationary objects (refer to Fig.~\ref{Case_study} (a) \textcolor{blue}{blue circles}). Consequently, it is arduous for such representation to depict dynamic actions.
To this end, we seek to answer the following question: \textit{Can we adaptively filter out repetitive events for stationary objects while preserving events recording the dynamic actions?}

Accordingly, we visualize the previous event frame representation~\cite{zheng2023deep} in Fig.~\ref{Case_study} (a). A salient observation is the `overlapped action region', as marked by the \textcolor{red}{red square}, which is a byproduct of the frame transformation process where consecutive actions overlap due to excessive event stacks. This ‘overlapped action region' thus serves as a pivotal indicator for inappropriate event sampling intervals. In light of this, we propose the AFE representation.


Specifically, to find the most appropriate dividing line of different actions based on the `overlapped action regions', we adopt the classic \textit{binary search} and implement it \textit{offline and recursively} with efficient $O(log{n})$ algorithm complexity. 
As illustrated in Fig.~\ref{Case_study} (b), the search algorithm can be seen as finding leaf nodes of a binary tree. To begin with, we equally divide the event stream $E^{0}_{0}$ (root) into two sub-streams $E^{0}_{1}$ (node) and $E^{1}_{1}$ (node) and thus generate their corresponding event count images $I^{0}_{1}$ and $I^{1}_{1}$. 
Then, we subtract $I^{0}_{1}$ with $I^{1}_{1}$ to obtain the difference map. 
Next, to measure the proportion of overlapped actions based on the difference map, we define a factor named deference rate $R$, given by: 
\begin{equation}
    R = sum(abs(I^{0}_{1} - I^{1}_{1})) / (sum(E^{0}_{0})/2),
\end{equation}
where the $sum(.)$ and $abs(.)$ functions indicate the event counts and absolute value operations, respectively. Intuitively, the high value of $R$ indicates a high probability of stacking events recording two different actions into one frame. In this case, we need to divide the event sub-stream recursively.

For the recursive algorithm, the boundary conditions are important. In our cases, if the difference rate $R$ is higher than the lowest sampling threshold $\Delta$, we repeat the above division process until it's lower than $\Delta$ or the event count number of the sub-stream $N_{sub}$ is less than the minimum aggregating event count number $N_{min}$. Note that hyper-parameters $N_{min}$ and $\Delta$ vary with different datasets. \textit{More discussions about selecting $N_{min}$ and $\Delta$ refer to the Supplmat.}
After the above searching process with $M$ times recursion, we finally obtain a series of fine-grained event frames  $I_{M}^{T}$ with $T$ event frames (all leaf nodes). 


\vspace{-2pt}
\subsection{Feature Encoding}
\vspace{-2pt}
\label{sec:feature_embedding}
With the AFE representation, the event stream is processed into a series of fine-grained event frames $I_{M}^{T}$.
Then, the event encoder and text encoder from the pre-trained event-text model of ~\cite{zhou2023clip} are utilized to establish a unified event-text embedding space.

\noindent\textbf{Event Encoder.} As shown in Fig.~\ref{Framework}, it inputs an RGB event frame $I_{M}^{i} \in \mathbb{R}^{H \times W \times 3}, i=1,2,...,T$ of the spatial size $H \times W$ and outputs the event embedding $f_{e}^{i}$. Given $T$ event frames, the event encoder processes $T$ times and generates the event [\textit{CLS}] embeddings $f_{e} \in \mathbb{R}^{T \times H_{e} \times W_{e} \times N_{e}}$.

\noindent\textbf{Text Encoder.} It takes two different kinds of text prompts as input: \textbf{1)} the hand-crafted text prompt {\fontfamily{qcr}\selectfont‘A series of photos recording action for} [\textit{CLASS}]{\fontfamily{qcr}\selectfont.'}, where [\textit{CLASS}] represents the category name. After encoding, each word is converted into the $D_{p}$-dimension word embedding and formed into the final text token $P_{h}$ 
\textbf{2)} the learnable text prompt $P_{l} = [P_{1}, P_{2},..., P_{n}, P_{CLASS}.]$, where $P_{i}, i=1,2,...,n_{l}$, is a random initialized parameter with $D_{p}$ dimensions; $n_{l}$ denotes the number of the learnable text prompts; $P_{CLASS}$ represents the encoded word embeddings of [\textit{CLASS}] and $[.]$ means the concatenation operation. Then, the text encoder transforms hand-crafted text prompt $P_{h}$ and learnable text prompt $P_{l}$ into corresponding text embeddings $f_{t}^{h}$ and $f_{t}^{l}$. We finally obtain the text [\textit{CLS}] embeddings $f_{t}$ 
by averaging $f_{t}^{h}$ and $f_{t}^{l}$.

\begin{figure}[t!]
\centering
\includegraphics[width=0.9\columnwidth]{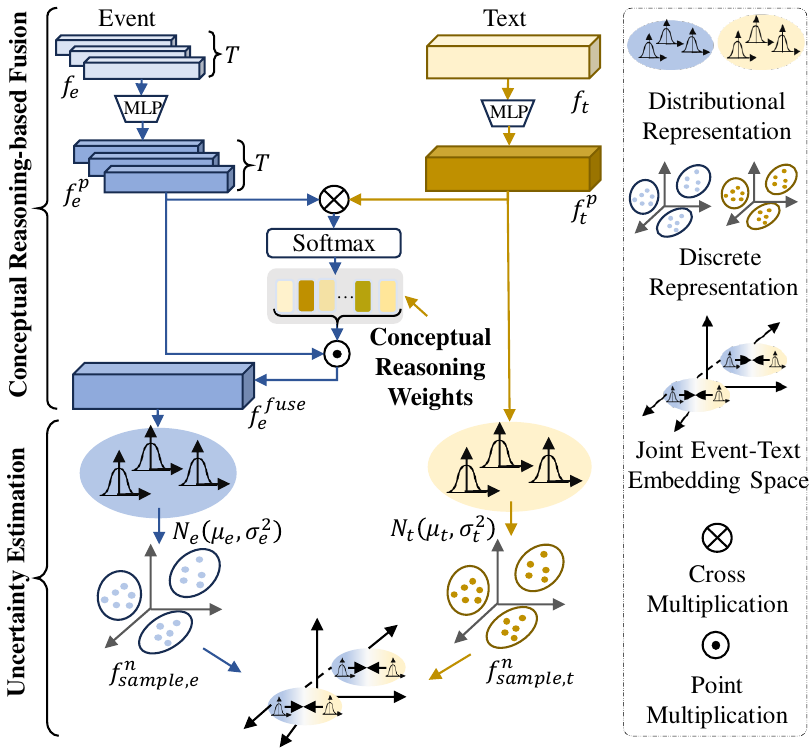}
\vspace{-2pt}
\caption{\textbf{The proposed CRUE module} consists of 1) conceptual reasoning for frame fusion based on the temporal relation among events and 2) uncertainty estimation of sub-actions for both text and event embeddings utilizing distributional representation. }\label{CRUE}
\vspace{-12pt}
\end{figure}

\vspace{-2pt}
\subsection{CRUE Module} 
\label{sec:CRUE}
\vspace{-2pt}
\label{Conceptualizing}
Previous event-based action recognition methods~\cite{lin2019tsm, gao2023action} fail to consider the following aspects:
\textbf{1)} \textbf{Temporal Relation}: Unlike stationary objects, dynamic actions unfold over time. The event data's temporal information is vital for understanding the meaning of an action. For instance, in Fig.~\ref{Teaser}, {\fontfamily{qcr}\selectfont‘Sit down'} and {\fontfamily{qcr}\selectfont‘Stand up'} involve similar sub-actions with the converse temporal occurrence, namely {\fontfamily{qcr}\selectfont‘Standing'} $\rightarrow$ {\fontfamily{qcr}\selectfont‘Squatting'} $\rightarrow$ {\fontfamily{qcr}\selectfont‘Sitting'} \textit{vs.} {\fontfamily{qcr}\selectfont‘Sitting'} $\rightarrow$ {\fontfamily{qcr}\selectfont‘Squatting'} $\rightarrow$ {\fontfamily{qcr}\selectfont‘Standing'}, thus resulting in different semantics.
\textbf{2)} \textbf{Semantic Uncertainty}: Actions, comprising various sub-actions, present greater semantic complexity than that of stationary objects. Take {\fontfamily{qcr}\selectfont‘Sit down'} as an example. It includes sub-actions: {\fontfamily{qcr}\selectfont‘Standing'}, {\fontfamily{qcr}\selectfont‘Squatting'}, and {\fontfamily{qcr}\selectfont‘Sitting'}. Each sub-action owns a specific meaning. Thus, using any sub-action is \textit{insufficient} and \textit{uncertain} to express the meaning of entire action {\fontfamily{qcr}\selectfont‘Sit down'}.  
Motivated by these two aspects, we propose the CRUE module to emulate the action recognition process of humans through the proposed conceptual reasoning-based fusion and uncertainty estimation.

\noindent\textbf{Conceptual Reasoning-based Fusion:} Unlike the simple average fusion of event frames~\cite{gao2023action,sabater2022event, xie2023event,plizzari2022e2}, the proposed CRUE module leverages the text embeddings to guide the event frame fusion. Specifically, as depicted in Fig.~\ref{CRUE}, given the event [\textit{CLS}] embeddings $f_{e} $ and text [\textit{CLS}] embeddings $f_{t}$, a two-layer MLP network is employed for dimension projection. In this way, we can obtain the projected event embeddings $f_{e}^{p}$ and text embeddings $f_{t}^{p}$. 
Then, we multiply $f_{e}^{p}$ and $f_{t}^{p}$ followed by the softmax function to generate the conceptual reasoning weights. Next, the conceptual reasoning weights are multiplied with the original projected event embeddings $f_{e}^{p}$ to obtain the fused event embeddings $f_{e}^{fuse}$. Intuitively, the conceptual reasoning weights are used as semantic weights generated based on event frames' temporal sequence for the frame fusion.

\noindent\textbf{Uncertainty Estimation:}
Semantic uncertainty refers to the obtained messages that tend to present multiple targets~\cite{ji2023map}. To model the semantic uncertainty of actions, we borrow the idea of distributional representation applied both in Natural Language Processing (NLP)~\cite{vilnis2014word,athiwaratkun2017multimodal,dasgupta2020improving} and Computer Vision (CV)~\cite{chang2020data,su2021self,sun2020view}. Unlike the methods that extract features as the discrete representation, we utilize the probability distribution encoder~\cite{ji2023map} for events and text embedding. 
Thus, the semantic uncertainty can be quantified by the variance of the probability distribution.

Specifically, as shown in Fig.~\ref{CRUE}, the fused event [\textit{CLS}] embeddings $f_{e}^{fuse}\in \mathbb{R}^{H_{e} \times W_{e} \times N_{e}}$ are equally split into $f_{e,1}^{fuse}$ and $f_{e,2}^{fuse}$ at the channel dimension~\cite{ji2023map}. Then, $f_{e,1}^{fuse}$ and $f_{e,2}^{fuse}$ are fed into two standard self-attention modules ~\cite{vaswani2017attention}. Next, we can predict a mean vector $\mu_{e} \in \mathbb{R}^{H_{e} \times W_{e} \times N_{e}}$ and a variance vector $\sigma_{e} \in \mathbb{R}^{H_{e} \times W_{e} \times N_{e}}$. Here, $\mu_{e}$ and $\sigma_{e}$ are the estimated parameters of the multivariate Gaussian distribution $f_{e}^{fuse} \sim N_{e}(\mu_{e},\sigma^{2}_{e})$. We conduct the same operation on the text \textit{CLS} embeddings $f_{t}$ to estimate its corresponding multivariate Gaussian distribution $f_{t}^{p} \sim N_{t}(\mu_{t},\sigma^{2}_{t})$. Overall, the above operations can be formulated as follows:
\begin{align}
  \mu_{i} = Att_{1}(f_{i,1}) ,\sigma_{i} = Att_{2}(f_{i,2}), \\
    f_{i} = [f_{i,2},f_{i,1}] \sim N_{i}(\mu_{i},\sigma^{2}_{i}),
\end{align}
where $i=e,t$ denotes the event and text embeddings respectively; $Att$ represents the self-attention module and $[.]$ indicates the concatenation operation.

With estimated distributional representation $N_{i}(\mu_{i},\sigma^{2}_{i}),$ $i=t,e$, we now quantify the semantic uncertainty of event and text embeddings. Then, we sample arbitrary discrete representation by employing the re-parameterization method~\cite{kingma2013auto} to ensure smooth back-propagation. That is, we first sample a random noise $\delta\sim N(0,I)$ from the standard normal distribution rather than sample directly from $f_{i} \sim N_{i}(\mu_{i},\sigma^{2}_{i}), i=t,e$. Next, we obtain the sampled discrete embeddings $f_{sample,i}^{n}$ by $f_{sample,i}^{n} = \mu_{i} + \delta\sigma_{i}$, where $n=1,2,..., N$ and $N$ is a hyper-parameter denoting the number of sampled discrete embeddings. (More discussion about $N$ refer to Sec.~\ref{ab_sample}.) 
The obtained $f_{sample,i}^{n}$ observes the estimated distribution $N_{i}$, and thus it can be used for estimating semantic uncertainty. 

However, the above random sampling process increases the complexity of training, especially given the spatial sparsity of event data. To accelerate model convergence, we introduce the \textit{Smooth L1} loss~\cite{girshick2015fast} between the normalized \textit{CLS} embeddings $f_{i}, i=t,e$ and the sampled embeddings $f_{sample, i}^{n}, i=t,e$:
\begin{small}
\begin{align} \label{loss_smoothL1}
   L_{smoothL1} = Smooth L1(f_{sample,i}^{n}, \frac {f_{i}-mean(f_{i})}{std(f_{i})}),
\end{align}
\end{small}
where $n=1,2,..., N$ and $N$ is a hyper-parameter denoting the number of sampled discrete embeddings, $mean(.)$ and $std(.)$ denote the mean and variance of the input embeddings. Besides, to reduce the semantic uncertainty of distributional representation, we introduce regularization loss:
\begin{align}\label{loss_reg}
   L_{reg} = sum(\sigma^{2}_{i}) + sum(\sigma^{2}_{i}), i=t,e,
\end{align}
where $sum(.)$ and $abs(.)$ indicate the summation of input embeddings and absolute value operation, respectively. Note that the experiment results show the final $L_{reg}$ is higher than zero when the model converges. This indicates the model doesn't degenerate from distributional representation into discrete representation as the variances are greater than zero.



\vspace{-3pt}
\subsection{Training Objectives}
To establish a joint event-text representation space for action recognition, we utilize the contrastive loss $L(f^{1}_{b}, f^{2}_{b})$ between two modal embeddings $f^{1}_{b}$ and $f^{2}_{b}$ as follows: 
\begin{small}
\begin{align}\label{loss_contrastive}
&L_{contrastive} (f^{1}_{b}, f^{2}_{b}) \nonumber
\\
&= 
-\frac{1}{B} \sum_{b \in B} \log \frac{\exp \left(f^{1}_{b} \times f^{2}_{b}/ \tau\right)}{\exp \left(f^{1}_{b} \times  f^{2}_{b} / \tau\right)+\sum_{b \neq \overline{b}} \exp \left(f^{1}_{b} \times f^{2}_{\overline{b}} / \tau\right)},
\end{align}
\end{small}
\vspace{-3pt}
where $\tau$ is the temperature coefficient, $B$ represents the size of the mini-batch, $b$ and $\overline{b}$ denote the $b$-th and the $\overline{b}$-th data among the mini-batch. 

We calculate the contrastive loss among all sampled event embeddings $f_{sample,e}^{n}$ and text embeddings $f_{sample,t}^{n}$. Finally, the whole training objective is composed of the contrastive loss, the \textit{Smooth L1} loss, and regularization loss combined with different rate hyperparameters:
\begin{align}
L_{final} =  & \alpha \times L_{contrastive}(f_{sample,e}^{n}, f_{sample,t}^{n}) \nonumber \\
&+ \beta \times L_{smoothL1} + \theta \times  L_{reg},
\end{align}
where we set the default values of $\alpha$, $\beta$ and $\theta$ to 1 after considering their numerical range.

\vspace{-3pt}
\subsection{SeAct Dataset}
\label{sec:SemanticAct}
\begin{figure}[t]
\centering
\includegraphics[width=1.0\columnwidth]{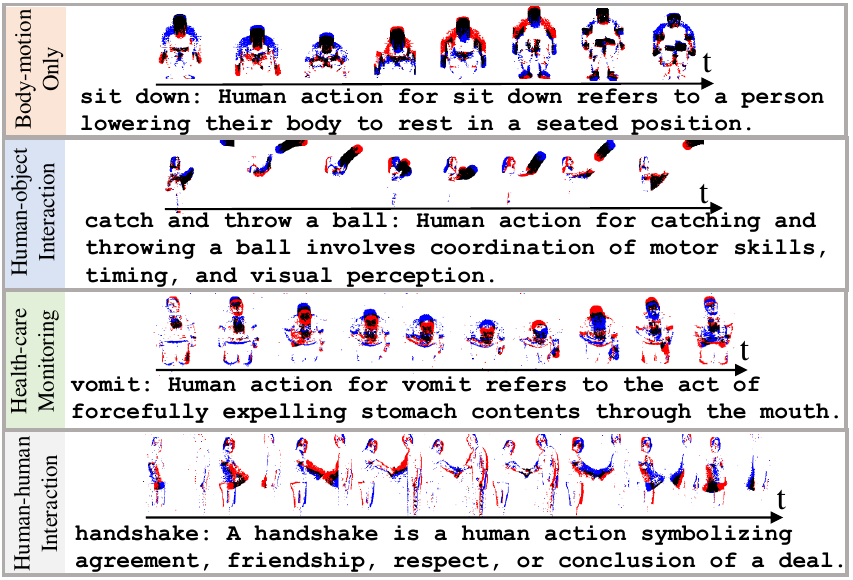}
\vspace{-15pt}
\caption{Examples of our SeAct dataset.}\label{dataset}
\vspace{-12pt}
\end{figure}

Since the category-level labels provided in the previous event action dataset~\cite{miao2019neuromorphic, wang2022hardvs, amir2017low, bi2020graph} only use several words to describe each action, they fail to stimulate the ability of our ExACT framework to process complex language information. To this end, we propose the \textbf{first} semantic-abundant SeAct dataset for event-text action recognition, where the detailed caption-level label of each action is provided. SeAct is collected with a DAVIS346 event camera whose resolution is 346 × 260. It contains 58 actions under four themes, as presented in Fig.~\ref{dataset}. Each action is accompanied by an action caption of less than 30 words generated by GPT-4~\cite{ray2023chatgpt} to enrich the semantic space of the original action labels. We split 80\% and 20\% of each category for training and testing (validating), respectively. We believe our SeAct dataset will be a better evaluation platform for event-text action recognition and inspire more relevant research in the future. \textit{Please refer to the Supplmat. for more dataset introduction.}

\vspace{-5pt}
\section{Experiments}
\subsection{Dataset and Experimental Settings}
\noindent \textbf{Dataset} In this work, four datasets are adopted for the evaluation of our proposed model, including PAF~\cite{miao2019neuromorphic}, HARDVS~\cite{wang2022hardvs}, DVS128 Gesture~\cite{amir2017low} and our newly proposed SeAct. PAF~\cite{miao2019neuromorphic}, also named Action Recognition, is a dataset recorded indoors, containing ten action categories with 450 recordings. HARDVS~\cite{wang2022hardvs} is a recently released dataset for event-based action recognition, currently having the largest action classes, namely 107,646 recordings for 300 action categories. Both of the above two datasets have an average time duration of 5 seconds with 346 × 260 resolution~\cite{wang2022hardvs}. DVS128 Gesture~\cite{amir2017low} is collected using a DVS128 camera with 128 × 128 resolution, dividing into 11 hand and arm gestures. \textit{Refer to Sec.~\ref{sec:SemanticAct} for the introduction of our SeAct dataset.}

\noindent \textbf{Experimental Settings}
In the AFE representation, the lowest sampling rate $\Delta$ is set as 50\%, 40\%, and 40\% while the minimum aggregating event number $N_{min}$ is chosen at 100,000, 150,000, and 100,000 on the PAF, HARDVS our SeAct datasets, respectively. The event encoder and text encoder of the event-image-text model ECLIP~\cite{zhou2023clip} are utilized for feature encoding. The number of sampled discrete embeddings $N$ is set to 5 based on the hyperparameter search. The initial learning rates are set to $1e-5$ with Adam optimizer~\cite{kingma2014adam} and weight decay equal to $2e-4$. CosineAnnealingLR~\cite{loshchilov2016sgdr} learning rate schedule is employed with a minimum learning rate of 1e-6. Our model is trained for 100 epochs for PAF and SeAct datasets, and 25 epochs for HARDVS. \textit{Please refer to the Supplmat. for additional experimental settings.}

\begin{table}[t!]
\centering
\setlength{\tabcolsep}{9pt}
\resizebox{0.9\linewidth}{!}{
\begin{tabular}{cccc}
\midrule
\multirow{2}{*}{Dataset} & \multirow{2}{*}{Model} & \multicolumn{2}{c}{Accuracy (\%)} \\ \cmidrule{3-4} 
 &  & Top-1 & Top-5 \\ \midrule
\multirow{7}{*}{PAF} & HMAX SNN~\cite{xiao2019event} & 55.00 & - \\
 & STCA~\cite{gu2019stca} & 71.20 &  \\
 & Motion   SNN~\cite{liu2021event} & 78.10 & - \\
 & MST~\cite{wang2023masked} & 88.21 & - \\
 & Swin-T   (BN)~\cite{wang2023masked} & 90.14 & - \\
 & EV-ACT~\cite{gao2023action} & \underline{92.60} & - \\
 & \textbf{ExACT (Ours)} & \textbf{94.83} & \textbf{98.28} \\ \midrule
\multirow{11}{*}{HARDVS} & X3D~\cite{feichtenhofer2020x3d} & 45.82 & 52.33 \\
 & SlowFast~\cite{feichtenhofer2019slowfast} & 46.54 & 54.76 \\
 & ACTION-Net~\cite{wang2021action} & 46.85 & 56.19 \\
 & R2Plus1D~\cite{tran2018closer} & 49.06 & 56.43 \\
 & ResNet18~\cite{he2016deep} & 49.20 & 56.09 \\
 & TAM~\cite{liu2021tam} & 50.41 & 57.99 \\
 & C3D~\cite{tran2015learning} & 50.52 & 56.14 \\
 & ESTF~\cite{wang2022hardvs} & 51.22 & 57.53 \\
 & Video-SwinTrans~\cite{liu2022video} & 51.91 & 59.11 \\
 & TSM~\cite{lin2019tsm} & \underline{52.63} & \underline{60.56} \\
 & \textbf{ExACT (Ours)} & \textbf{90.10} & \textbf{96.69} \\ \midrule
\multirow{8}{*}{DVS128 Gesture} &  Time-surfaces~\cite{maro2020event} & 90.62 & - \\
& SNN eRBP~\cite{kaiser2019embodied} & 92.70 & - \\
& Slayer~\cite{shrestha2018slayer} & 93.64 & - \\
& DECOLLE~\cite{kaiser2020synaptic} & 95.54 & - \\
& EvT~\cite{sabater2022event} & 96.20 & - \\
& TBR~\cite{innocenti2021temporal} & 97.73 & - \\
& EventTransAct~\cite{de2023eventtransact} & \underline{97.92} & - \\
& \textbf{ExACT (Ours)} & \textbf{98.86} & \textbf{98.86} \\ \midrule
\multirow{4}{*}{SeAct} & EventTransAct~\cite{de2023eventtransact}  & 57.81 & 64.22 \\
& EvT~\cite{sabater2022event} & 61.30 & 67.81 \\
 & \textbf{ExACT-category} & \underline{66.07} & \underline{70.54} \\
 & \textbf{ExACT-caption} & \textbf{67.24} & \textbf{75.00} \\ \midrule
\end{tabular}}
\vspace{-5pt}
\caption{An overall comparison with SoTA models for event-based action recognition task on the PAF, HARDVS, DVS128 Gesture, and our proposed SeAct dataset. The best scores are in bold, and the second scores are underlined. `ExACT-category' and `ExACT-caption' represent that ExACT is trained with category-level and caption-level labels, respectively.}
\label{result_recognition}
\vspace{-13pt}
\end{table}

\vspace{-3pt}
\subsection{Comparison with SOTA Methods}
As shown in Tab.~\ref{result_recognition}, our proposed ExACT demonstrates superior performance on the PAF and HARDVS datasets. Specifically, ExACT brings +2.23\% improvements on the PAF dataset with ten classes compared with SOTA results. Notably, ExACT brings remarkable +37.47\% improvements on the HARDVS with 300 classes, showcasing ExACT's excellent potential in classifying complex and diverse actions. Moreover, upon evaluation using our proposed SeAct dataset,  ExACT exhibits a 67.24\% Top-1 and 75.00\% Top-5 recognition accuracy in real-world scenarios involving 58 dynamic actions with caption-level labels. These results demonstrate the effectiveness of our ExACT framework in action recognition.

\vspace{-2pt}
\subsection{Ablation Studies}
In this section, we ablate the key components of ExACT, training objectives, and important hyper-parameters to explore their effectiveness. Unless otherwise stated, experiments are performed on the PAF dataset.

\noindent\textbf{Effectiveness of the AFE representation.}
Tab.~\ref{result_representation} shows that, with a comparatively lower sampled number of 2816, our AFE representation obtains the best accuracy of 94.83\%.
This result indicates that our AFE representation can achieve better performance by filtering out the repetitive frames portraying the same action.
\begin{table}[t!]
\centering
\setlength{\tabcolsep}{8pt}
\resizebox{\linewidth}{!}{
\begin{tabular}{ccccc}
\midrule
\multirow{2}{*}{Sample strategy} & \multirow{2}{*}{Frame number} & \multirow{2}{*}{Aggregated method} & \multicolumn{2}{c}{Accuracy} \\ \cmidrule{4-5} 
 & & & Top-1 & Top-5 \\ \midrule
\multirow{3}{*}{Event histogram} & 3122 & every 80,000 events & 89.29 & 91.07 \\
 & 2720 & every 90,000 events & 94.21 & 97.14 \\
 & 2405 & every 100,000 events & 93.10 & 95.24 \\ \midrule
\multirow{3}{*}{Event voxel~\cite{deng2022voxel}} & 3122 & every 80,000 events & 88.88 & 90.46 \\
 & 2720 & every 90,000 events & 92.45 & 94.89 \\
 & 2405 & every 100,000 events & 90.85 & 92.31 \\\midrule
TBR~\cite{innocenti2021temporal} & 2758 & every 2000 ms events & 92.79 & 95.16\\ \midrule
AFE (Ours) & 2894 & adaptive events number & \textbf{94.83} & \textbf{98.28} \\ \midrule
\end{tabular}}
\vspace{-7pt}
\caption{The comparison of AFE representation with previous event frame representation.}\label{result_representation}
\vspace{-6pt}
\end{table}

\begin{table}[t]
\centering
\setlength{\tabcolsep}{14pt}
\resizebox{\linewidth}{!}{
\begin{tabular}{lcccc}
\midrule
\multirow{2}{*}{Method} & \multicolumn{4}{c}{Accuracy} \\ \cmidrule{2-5} 
 & Top-1 & $\Delta$ & Top-5 & $\Delta$ \\ \midrule
Contrastive & 92.86 & - & 94.64 & -  \\
+ CR  & 93.64 & \cellcolor{gray!10}+0.78& 96.43 & \cellcolor{gray!10}+1.79\\
+ CR, UE & \textbf{94.83} & \cellcolor{gray!20}+1.97& \textbf{98.28} & \cellcolor{gray!20}+3.64\\ \midrule
\end{tabular}}
\vspace{-7pt}
\caption{The ablation study of the CRUE module, where Contrastive, CR, and UE denote the contrastive learning loss function, proposed conceptual reasoning-based fusion and uncertainty estimation, respectively.}\label{ab_conceptualizing}
\vspace{-6pt}
\end{table}

\begin{table}[t]
\centering
\setlength{\tabcolsep}{14pt}
\resizebox{\linewidth}{!}{
\begin{tabular}{lcccc}
\midrule
\multicolumn{1}{c}{\multirow{2}{*}{Method}} & \multicolumn{4}{c}{Accuracy} \\ \cmidrule{2-5} 
\multicolumn{1}{c}{} & Top-1 & $\Delta$ & Top-5 &  $\Delta$\\ \midrule
Sum & 89.14 & - & 90.97 & - \\
Mean Pooling & 92.52 & \cellcolor{gray!10}+3.38 & 95.04 &\cellcolor{gray!10} +5.07 \\
CR & \textbf{94.83}&\cellcolor{gray!20} +5.69 & \textbf{98.28} &\cellcolor{gray!20} +8.31\\ \midrule
\end{tabular}}
\vspace{-7pt}
\caption{Impact of CR operation proposed in the CRUE module.}
\vspace{-12pt}
\end{table}

\noindent\textbf{CRUE module \textit{vs.} Contrastive learning.}
As shown in Tab.~\ref{ab_conceptualizing}, the baseline model is trained using the contrastive learning loss, which is widely adopted in previous methods~\cite{zhou2023clip,wu2023eventclip}. 
Results indicate that the Conceptual Reasoning-based fusion (CR) and CR with Uncertainty Estimation (UE) lead to an increase in accuracy of +0.78\% and +1.97\%, respectively. This implies that the CRUE module enhances the model's ability to comprehend actions by conceptual reasoning-based fusing event frames based on their temporal relations and estimating the uncertainty of action semantics during training, thus achieving better performance than simply employing contrastive learning.

\noindent\textbf{Conceptual Reasoning-based fusion (CR) \textit{vs.} Other frame fusion methods}
To evaluate the effectiveness of the CR. We compare the CR with two other designs, namely, the sum and average pooling of all event frames, as shown in Tab.~\ref{ab_conceptualizing}. The results demonstrate the effectiveness of our proposed CR, as it improves recognition accuracy by +5.69\% and +2.31\% when compared to the sum operation and the mean pooling operation, respectively.

\begin{table}[t]
\centering
\setlength{\tabcolsep}{14pt}
\resizebox{\linewidth}{!}{
\begin{tabular}{lcccc}
\midrule
\multicolumn{1}{c}{\multirow{2}{*}{Training Method}} & \multicolumn{4}{c}{Accuracy} \\ \cmidrule{2-5} 
\multicolumn{1}{c}{} & Top-1 & $\Delta$ & Top-5 & $\Delta$ \\ \midrule
$L_{contrastive}$ & 92.86 & -& 94.64 &- \\
+ $L_{reg}$ & 93.96 & \cellcolor{gray!10}+1.10 & 95.82 & +1.18\cellcolor{gray!10} \\
+ $L_{smoothL1}$ & 94.41 & \cellcolor{gray!20}+1.55& 97.89 &\cellcolor{gray!20}+3.25 \\
+ $L_{reg}$, $L_{smoothL1}$ & \textbf{94.83} &\cellcolor{gray!30}+1.97 & \textbf{98.28}&\cellcolor{gray!30}+3.46  \\ \midrule
\end{tabular}}
\vspace{-7pt}
\caption{Impact of different training objectives.}
\vspace{-5pt}
\label{ab_training_objectives}
\end{table}

\begin{figure}[t!]
\centering
\includegraphics[width=0.7\columnwidth]{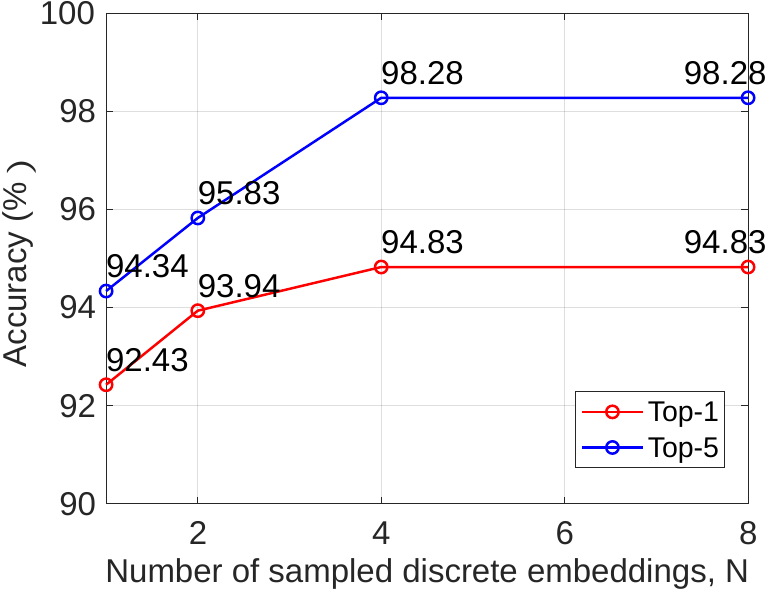}
\vspace{-7pt}
\caption{Impact of different numbers of sampled discrete embeddings proposed in the CRUE Module.}\label{ab_samplenumber}
\vspace{-12pt}
\end{figure}

\noindent\textbf{Performance comparison of different training objectives.}
Tab.~\ref{ab_training_objectives} shows the impact of different training objectives on model performance. Model trained using the contrastive learning loss $L_{contrastive}$ (Eq.~\ref{loss_contrastive}) exhibits the lowest performance compared to other combinations of training objectives. Both the $L_{reg}$ (Eq.~\ref{loss_reg}) and $L_{smoothL1}$ (Eq.~\ref{loss_smoothL1}) enhance the model performance, bringing +1.10\% and +1.55\% accuracy improvements respectively. The combination of all training objectives brings the largest performance improvement of +1.97\%, which further validates the effectiveness of the proposed CRUE module.

\noindent\textbf{Impact of the different number of point sampling.}
\label{ab_sample}
As shown in Fig.~\ref{ab_samplenumber}, we explore the effect of the number of sampled discrete embeddings $N$. We find that as $N$ increases from 1 to 4, the accuracy increases from 92.43\% to 94.83\%. When $N$ increases from 4 to 8, the accuracy remains constant, indicating that increasing $N$ yields diminishing performance accuracy improvement. Intuitively, this comes from the fact that distributional representation introduces disturbance during training while more sampled discrete embeddings mitigate this disturbance. Consequently, we set the hyperparameter $N$ as $5$ during training. 

\begin{figure}[t]
\centering
\includegraphics[width=1.0\columnwidth]{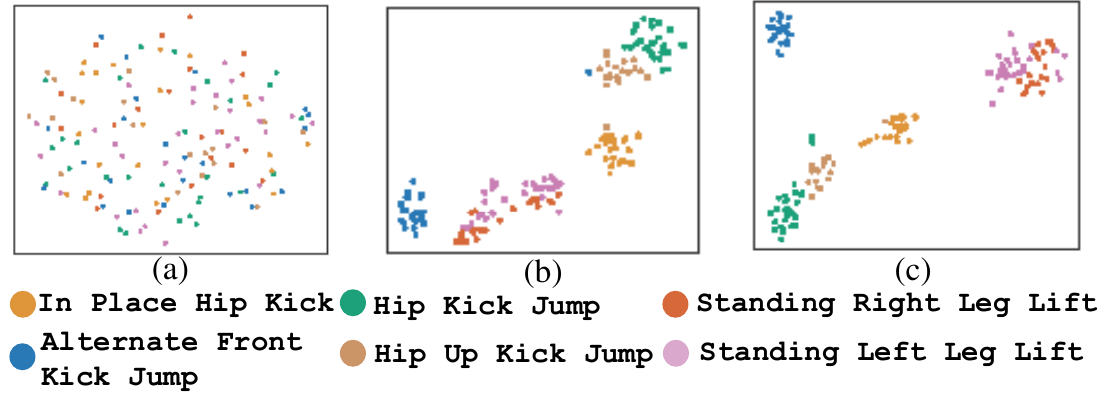}
\vspace{-15pt}
\caption{t-SNE visualization of event embeddings on the HARDVS dataset. (a) Before training; (b) Training without CRUE module; (c) Training with CRUE module.}\label{t-sne}
\vspace{-14pt}
\end{figure}

\noindent\textbf{The t-SNE visualization of event embeddings for the CRUE module.} We select 144 event examples belonging to six categories from the HARDVS dataset. These categories include a simple pair with different semantics, as well as two challenging pairs with similar semantics, namely {\fontfamily{qcr}\selectfont‘Hip Kick Jump'} \textit{vs.} {\fontfamily{qcr}\selectfont‘Hip Up Kick Jump'} and {\fontfamily{qcr}\selectfont‘Standing Right Leg Lift'} \textit{vs.} {\fontfamily{qcr}\selectfont‘"Standing Left Leg Lift"'}. Fig.~\ref{t-sne} displays the alteration of event embedding distribution before training and training with/without the CRUE module. Comparing those hard pairs in Fig.~\ref{t-sne} (b) and (c), we can find that event embeddings in different categories are widely distributed while our proposed CRUE module advances in distinguishing those hard pairs with similar semantic meanings. Intuitively, visualization results prove that the CRUE module enhances the model's performance by expressing the uncertainty of semantics, especially for those with similar semantics. \textit{See more results in Supplmat.}

\subsection{Extension to Other Tasks}
In this section, we transfer our EXACT model to both text-to-event and event-to-text retrieval tasks. Retrieval refers to searching and retrieving required data that matches a given query. The query can be of different modalities like events or texts. It has several practical applications, such as matching a real-world abnormal action with its corresponding event data in surveillance scenarios. For action retrieval using EXACT, we feed the query and events/texts into corresponding encoders to obtain their embeddings. Next, we calculate the similarity score between the query embedding and event/text embeddings, then select those events/texts with the highest similarity scores as the output. All retrieval experiments are conducted on our SeAct dataset. \textit{Refer to Supplmat. for more retrieving results.}

\begin{figure}[t]
\centering
\includegraphics[width=1.0\columnwidth]{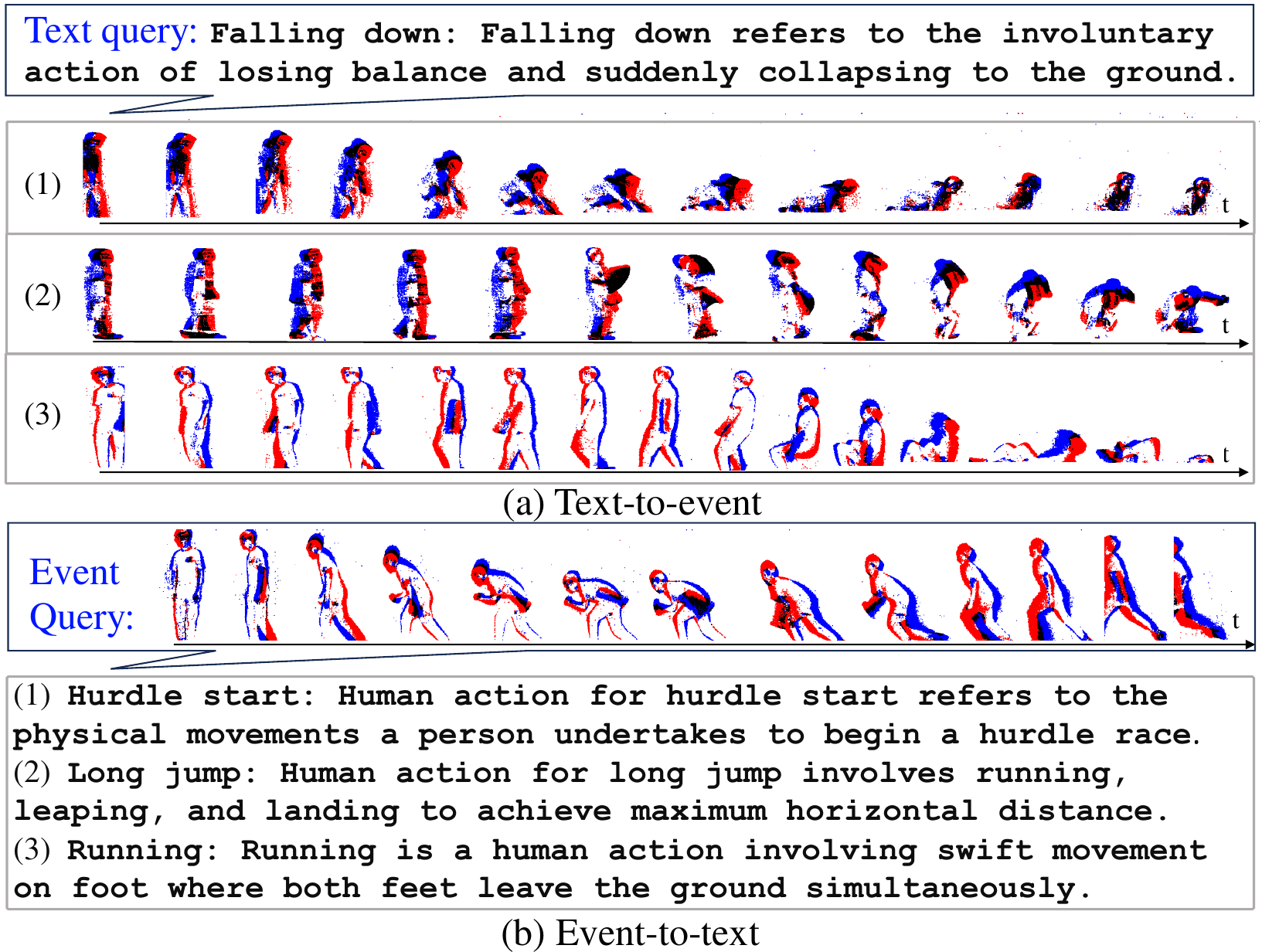}
\vspace{-18pt}
\caption{Action retrieval results.}\label{Event-to-text_retrival}
\vspace{-14pt}
\end{figure}

\noindent\textbf{Text-to-event action retrieval} In Fig.~\ref{Event-to-text_retrival} (a), we present the Top-3 retrieved event streams utilizing the action {\fontfamily{qcr}\selectfont‘Falling down'} with captions. All retrieved event data exhibit a high degree of similarity to the input text query, providing further evidence of EXACT's effectiveness.

\noindent\textbf{Event-to-text action retrieval} In Fig.~\ref{Event-to-text_retrival} (b), we display the Top-3 retrieved text captions queried by the event stream recording action {\fontfamily{qcr}\selectfont‘Hurdle start'}. The retrieved captions include actions of {\fontfamily{qcr}\selectfont‘Hurdle start'}, {\fontfamily{qcr}\selectfont‘Long jump'} and {\fontfamily{qcr}\selectfont‘Running'}, which share several sub-actions of similar semantics, proving the effectiveness of ExACT.

\vspace{-6pt}
\section{Conclusion and Future Work}
We presented the ExACT, as the first exploration utilizing language guidance for event-based action recognition. We proposed a CRUE module to simulate the action recognition of human beings, especially focusing on conceptual reasoning of the temporal relations among event frames and estimating the uncertainty of actions. Besides, we proposed the AFE representation, which adaptively eliminated repetitive events to generate detailed event frames for dynamic actions. To evaluate models’ understanding of the complex semantics of actions involving multiple sub-actions with different semantics, we presented the SeAct dataset with semantic-abundant action captions as the first benchmark for event-text action recognition. Our ExACT framework achieved SOTA results on the PAF and HARDVA datasets and achieved plausible performance on our SeAct dataset. Furthermore, we extended ExACT to event-text retrieval tasks, proving its flexible transferability. 

\noindent\textbf{Future Work:} In the future, we will enhance the conceptual reasoning and uncertainty estimation module in various event vision tasks and experiments on larger event action datasets with semantic-abundant caption-level labels.

\noindent\textbf{Acknowledgement}
This paper is supported by the National Natural Science Foundation of China (NSF) under Grant No. NSFC22FYT45 and the Guangzhou City, University and Enterprise Joint Fund under Grant No.SL2022A03J01278.
\clearpage
{
    \small
    \bibliographystyle{ieeenat_fullname}
    \bibliography{main}
}


\end{document}